\documentclass{article}
\usepackage{ijcai17}

\usepackage{times}

\usepackage{graphicx} 

\usepackage{natbib}

\usepackage{algorithm}
\usepackage{algorithmic}

\usepackage{hyperref}

\usepackage[caption=false,font=normalsize,labelfont=sf,textfont=sf]{subfig}

\usepackage{url}

\title{The Case for Meta-Cognitive Machine Learning:\\On Model Entropy and Concept Formation in Deep Learning}
\author{Johan Loeckx\\ Artificial Intelligence Lab, Vrije Universiteit Brussel, Pleinlaan 2, 1050 Brussel \\ jloeckx@ai.vub.ac.be}

\begin{document}
\maketitle

\begin{abstract}
Machine learning is usually defined in behaviourist terms, where external validation is the primary mechanism of learning.  In this paper, I argue for a more holistic interpretation in which finding more probable, efficient and abstract representations is as central to learning as performance.  In other words, machine learning should be extended with strategies to reason over its own learning process, leading to so-called meta-cognitive machine learning. As such, the de facto definition of machine learning should be reformulated in these intrinsically multi-objective terms, taking into account not only the task performance but also internal learning objectives. To this end, we suggest a ``model entropy function" to be defined that quantifies the efficiency of the internal learning processes.  It is conjured that the minimization of this model entropy leads to concept formation. Besides philosophical aspects, some initial illustrations are included to support the claims.
\end{abstract}

\section{Introduction}
Machine learning is often approached from a behaviourist perspective, in which external feedback in the form of a reinforcement signal is the major driving force of improvement. 
Though this method has lead to many successes, it is confronted with interesting and unsolved challenges like tackling overfitting, providing comprehensibility, building reusable abstractions and concept formation, among many other \citep{kotsiantis2007supervised, bengio2009learning}. 
The problem with these behaviourist approaches is that they ignore the central importance of internal processes when considering learning. Model internals are often regarded just as a \textit{means} to achieve higher performance.  Analogous to studying human behaviour, however, appreciating the mechanisms of learning boils down to the question: "when have we really learnt?" 
%
In this paper, we argue that a computer has learnt when:
\begin{itemize}
\item the programme becomes better at the task at hand; 
\item the programme can perform the task more efficiently;
\item the code becomes "more structured" and simpler.
\end{itemize}
One possible analogy to better understand the above statements can be found in software engineering.  When considering code that performs a specific task, we do not care only about its functionality, but also about its execution speed/efficiency and other so-called "non-functional requirements".   Furthermore, a carefully modularized design probably reflects more understanding than an endless enumeration of IF-ELSE clauses.
\begin{quote}
\textbf{In other words, finding a more efficient and structured way to represent/reproduce information and to perform a learning task, \textit{is as central to machine learning as the reproduction of results.}}
\end{quote}
Different to humans, of course, machines are measurable.  This provides us with a unique opportunity to study the nature of learning in principle, at the same time improving Machine Intelligence.
We are not claiming that model complexity/efficiency has not been subject to past research efforts.  On the contrary, many techniques and design principles have attempted to improve exactly these properties -- like Occam's razor, Bayesian structure learning, pruning, the use of prototypes to compact information, regularization as a strategy to reduce energy, weight sharing in RNNs or CNNs to decrease model complexity, etc.

Indeed, the whole evolution of Deep Learning can be seen as one specific approach in the quest to find models that are more structured (i.e. have a lower entropy), by organizing and training them in a layer-wise fashion  \citep{bengio2009learning}.  
The focus has been mainly on training algorithms and designing model architectures that are adapted to these kinds of ``deep" structures \citep{deng2014deep}.  Similar to efforts in multi-objective machine learning, these techniques are considered as a means to improve (externally measured) performance rather than a goal in itself \citep{jin2008pareto}.
We, however, do believe that minimizing the model's structural complexity and optimizing its efficiency of representation, is not only a \textit{means} to improve (externally validated) performance, but a central pillar to machine intelligence that leads to concept formulation and should be made explicit.  
In this sense, our vision aligns to that of Ray Kurzweil, who claimed that ``the theory behind deep learning\ldots is that you have a model that \textit{reflects the hierarchy in the natural phenomenon you're trying to learn} \citep{kurzweil2013}."

This paper is structured as follows.  The theoretical ideas are laid out and the case for a new operational definition of machine learning is made.  We put forward the conjecture that the optimization of model entropy, leads to concept formation. 
Last, conclusions and further steps to operationalize these concepts are formulated.


\begin{figure}[!t]
\centering
\includegraphics[width=0.30\textwidth]{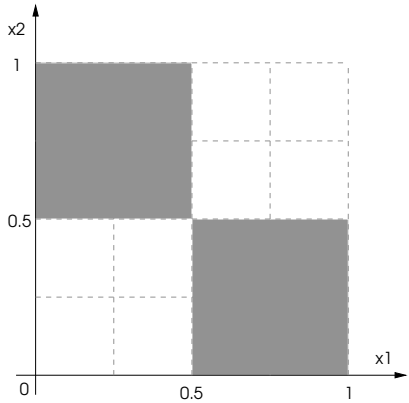}
\caption{Very basic example of a concept to be learnt.  The XOR function above represents an anti-symmetric
function in two variables x1 and x2.  We expect a good model to reflect this anti-symmetry in the model structure or parameters.} 
\label{fig:xor-problem}
\end{figure}

\section{Central assertions}
\subsection{Learning can not be explained in extrinsic terms only}
Conventional wisdom depicts machine learning as the optimization of a (non-)parametric model with respect to some performance measure. This view is clearly reflected in the de facto definition of machine learning by Mitchell \citep{mitchell1997machine}:  
"A computer program is said to learn from an experience $X$ with respect to some class of tasks $T$ and performance measure $P$, if its performance at tasks in $T$, as measured by $P$, improves with experimental data $D$".

Traditional machine learning techniques typically exploit shallow-structured, and often fixed, architectures.   Nevertheless, there is a general consensus that the learning of ``higher-order" concepts is problematic, and that the solution to this issue is somehow connected to deep architectures that create ever higher forms of abstraction.  Experimental research as well as neurological evidence on the organization of the brain, supports this finding \citep{bianchini2014complexity}. The limitation of architecture complexity is preferred, primarily because their behaviour could be understood and the training of more complex or adaptive architectures leads to a explosion of complexity.

That was until recently.  The recent advanced in so-called ``Deep Learning", have focused on training algorithms that are adapted to new kinds of deep architectures \citep{deng2014deep}, and heuristic strategies to attain specific structural properties like sparse coding that lead to higher forms of abstractions. With the exception of studies on ``interpretability" \citep{jin2008pareto}, structural properties are mainly considered a by-product, a (desirable) side effect of the applied training mechanisms. 
Though the organization and complexity of model topologies is acknowledged to be crucial, current approaches are mainly limited to analysing the data space, i.e. the implemented regression functions or decision boundaries \citep{bianchini2014complexity}.

\begin{figure*}[!t]
\centering
\subfloat[Minimal entropy model]{\includegraphics[width=0.42\textwidth]{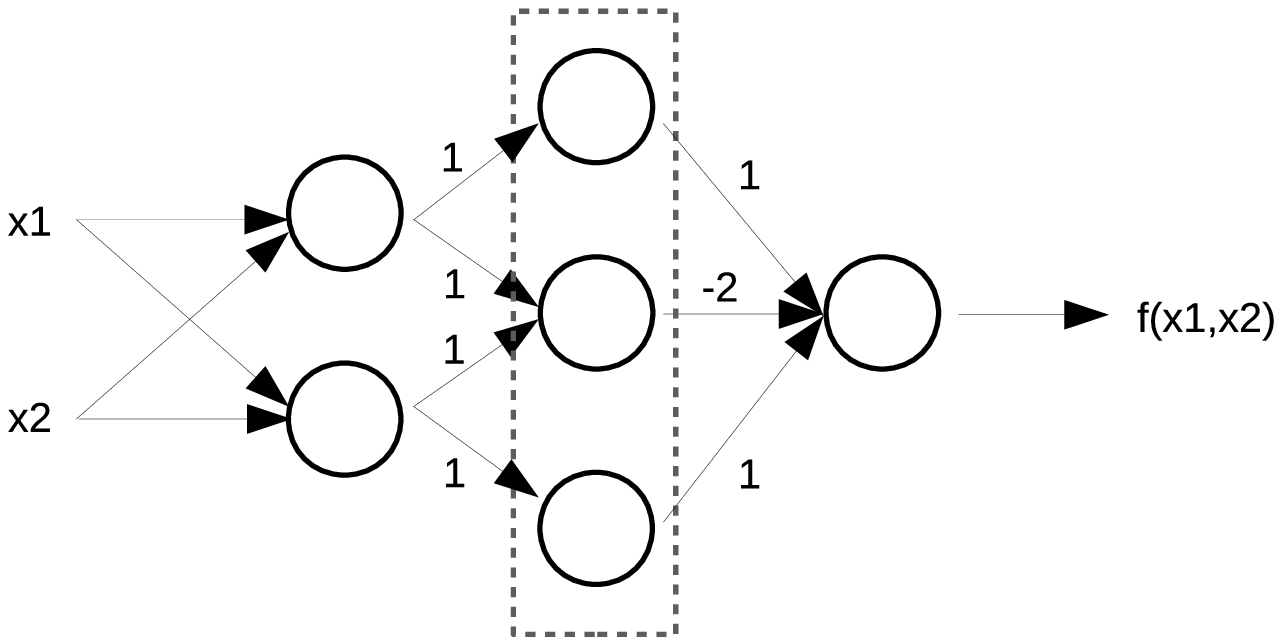}
\label{fig:arch_minS}}
\hfil
\subfloat[Non-optimal entropy model]{\includegraphics[width=0.4\textwidth]{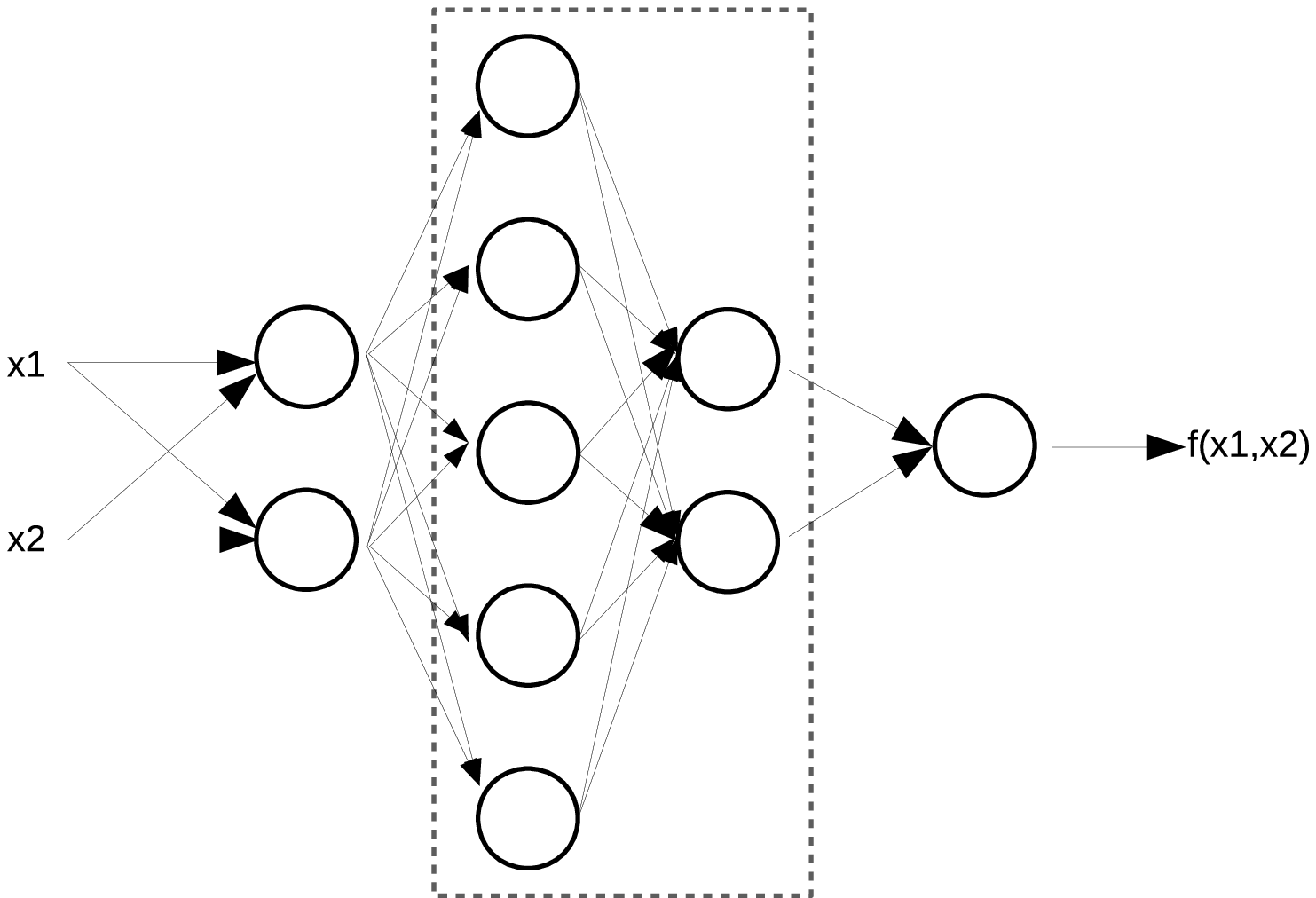}
\label{fig:arch_medS}}
\caption{Model internals matter. Example zero-test-error neural network topologies that characterize an XOR gate. The inputs and outputs were defined on the domain \{0,1\}. Figure (a) shows a minimal entropy model, (b) a non-minimal entropy one.  Both networks have an identical decision surface, and therefore identical generalization and noise-robustness properties.  However, the surfaces were constructed from substantially different model structures. }
\label{fig:xor_architectures}
\end{figure*}

There is a problem with this approach.  Consider an neural network algorithm that needs to learn a simple concept like an ``XOR" function depicted in Fig. \ref{fig:xor-problem}.  An infinite number of neural networks with very similar or identical decision boundaries can be constructed -- of which two are shown in Fig. \ref{fig:xor_architectures}.  From an external point of view, there is no way to discriminate between these two models: describing the difference between these two models can only occur in terms of the model internals.  
Of course the weight space, which represents the model of a neural network, is related to the data space, as it performs calculations on the data.  In other words:
\begin{quote}
\textbf{Data representation and model computation should be considered as two sides of the same coin. As a result the structural properties of \textit{both the model and data space} are key to the modelling of higher abstractions.}
\end{quote}
Sparse coding is a perfect example of this.  Without sparse coding, although the information is intrinsically ``present" in the data, neural networks become intractable to train due to the extremely volatile and complex decision surface.  
From this perspective we follow the observations that have been made by Bengio in \citep{bengio2013representation} on representation learning.

One of the interesting phenomena is ``information entanglement'' \citep{glorot2011deep}.  In this case, the model space is of a lower dimensionality or complexity than the data space.   The projection of the data onto a high-dimensional space using sparse coding, then, has the advantage that the representations are more likely to be linearly separable, or at least less nonlinear.  On the other hand, when the model complexity is increased considerably (e.g. by adding layers), the neural network becomes untrainable using traditional techniques, because the dimensionality of the search space explodes.    Deep learning techniques tackle this issue by -- among other techniques -- pre-initializing the model-space of particular layer in a maximum-likelihood/minimal-energy state. 


\subsection{Machine learning regards strategies that optimize performance, model structure and data modelling efficiency/likelihood}
Though the energy-based approach has proven very successful, the choice to divide and train a network layer-wise, is rather arbitrary.   Instead, we would like to have some kind of process that optimizes the model structure according to the data and leads to representation learning. For example, by driving weights to stimulate the formation of sparse representations and promote disentanglement. In this vision, learning by machines transforms into the implementation and execution of  \textbf{strategies} that optimize, not only the external performance and inner energetic properties of the model, but as well the model entropy.  In other words, 
\begin{quote}
\textbf{Machine learning is an intrinsically multi-objective process that trades off \textit{task performance} for \textit{model complexity} and data modelling efficiency/energy or \textit{model likelihood}.}
\end{quote}
There are many advantages to moving away from a purely ``extrinsic'' definition of Machine Learning towards a  cognitively inspired operational definition \citep{turing1950computing}.  To start, the artificial \textit{distinction between supervised and unsupervised learning evaporates} as a particular model can still being optimized, but towards minimal entropy rather than minimal error. This also opens the door towards learning from few symbolic data.  The multi-objective approach allows a more deep comparison between different machine learning models, and even between different kinds of algorithms.   It allows a more focused approach towards deep learning techniques and lastly, it opens up new opportunities to quantize biases in machine learning algorithms.
\begin{quote}
\textbf{(Machine) Learning implements \textit{strategies} to optimize performance as well as model likelihood and entropy.}
\end{quote}
We will now briefly discuss these three criteria or objectives.

\subsubsection{Extrinsic performance (P)}  P determines how well a model scores when performing an externally defined task.  Often not only the raw performance counts, but also additional desirable properties like noise-immunity, smoothness, generalization to unseen data, etc.  These qualities can only be measured in terms of the externally defined problem.

\subsubsection{Data modelling efficiency (E)} A model that requires less energy to represent the data to be "learnt", probably demonstrates superior insight.   Modelling efficiency can thus be interpreted as the \textit{likelihood that a particular model has generated particular data}.  The optimization of the margin in Support Vector Machines is an example. Maximum likelihood models are related to ``maximal entropy models". These models attempt to minimize the a-priori assumptions imposed by the model on the \emph{data}.  Minimal entropy models and maximal entropy modelling are not contradictory nor similar.  In the latter case, entropy is measured in the \textit{data space} (the \textit{output} of the model), and not in the \textit{meta-space} of model parameters (e.g. weights).  
Especially in the presence of uncertainty (for example in the case of noise or sub-sampling), data modelling efficiency is essential to improve the robustness of a model.  We will elaborate on this in later sections.

\subsubsection{Model entropy (S)}
Considering model complexity dates back to the beginning of computational modelling \citep{barron1991complexity, bartlett1998sample, myung2000importance, spiegelhalter2002bayesian, kon2006complexity}. It makes indeed sense to believe that models that have a lower entropy or complexity (e.g. by exploiting hierarchy, symmetry or other regularities) without losing expressive power, provide superior abstraction.  It therefore makes sense to optimize model structures for \textit{minimal} entropy with respect to the data.  Model entropy is determined not only by the network topology, but also the value and patterns in the weights values and their interaction with data (more on this later).  

It is tempting to confuse minimizing model entropy with methods assuming ``smoothness" of the input/output data distribute or impose structural constraints on the \textit{data}.  While they are clearly connected, there is a crucial difference. While minimal entropy models elevate the abstraction of a model for a given set of data, they do not assume a low \emph{data entropy} as is the case for example in the methods referred to by \citep{bengio2009learning}.

\subsection{The definition of (machine) learning needs to be extended} 
A lot of algorithms already implicitly employ these kind of optimization strategies. The usage of prototypes or nearest-neighbours are strategies to compact information and implicitly improve modelling efficiency as the class of a sample is determined by the distance to the closest neighbour; regularization a strategy to reducing energy consumption; weight sharing to decrease the complexity/entropy of a model; penalty functions to constrain model complexity; Occam's razor or pruning in decision trees.

However, to acknowledge the equal importance of all three aspects, the definition of machine learning should be extended:
\begin{quote}
\textbf{``A computer program is said to learn from experience $X$ if its performance at task in T, as measured by P, improves with experience $X$ or when the model entropy $S$ or its modelling efficiency or model likelihood $E$ decreases with respect to the training data $D$ associated with experience $X$''.}
\end{quote}

\begin{figure*}[!t]
\centering
\subfloat[Problem stated in a Cartesian coordinate system.]{\includegraphics[width=0.32\textwidth]{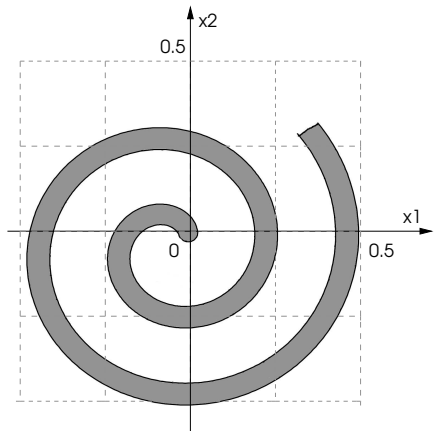}%
\label{fig:spiral-cartesian}}
\hfil
\subfloat[Problem stated in a polar coordinate system.]
{\includegraphics[width=0.33\textwidth]{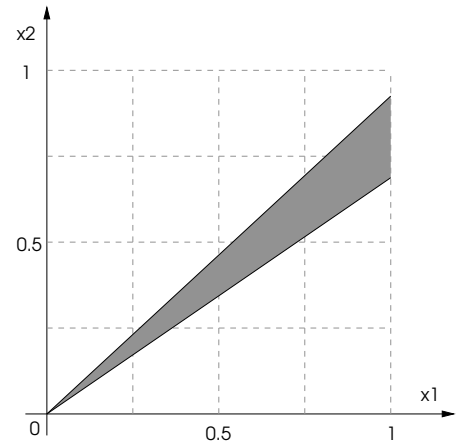}%
\label{fig:spiral-polar}}
\caption{The same ``double spiral'' learning problem starting from different input data abstraction (qualitative view).  The complexity of a model depends on properties of both model as data.  Computation and representations are two sides of the same coin.} 
\label{fig:twospirals}
\end{figure*}

\subsection{An ``entropy function" needs to be defined}
Similar to the ``energy function'' in traditional deep learning, an entropy function should be defined that measures the complexity of the \textit{process} to construct higher-order abstract representations and represents the efficiency of the learning processes.  Existing research has explored similar ideas,  in which, for example, neural networks are characterized by the impact of the topology on the reduction of information theoretic content through the network \citep{tishby2015deep}.

However, it is crucial to realize that this complexity depends on \textit{both} \textit{model characteristics} like topology and weight patterns, and properties of \textit{sample configurations} (a configuration is the combined representation of all internal and external signals, for example the outputs of all neurons). Though one could be tempted to think the entropy \textit{fixed} for a given set of training data $D$, this is not the case in the process of abstraction. Indeed, new, more abstract representations are produced from old ones using computation and thus the complexity of this calculation matters.  On the other side, the complexity of this computation depends on the data representation as well.  This, of course, is in line with previous investigations into the nature of complexity \citep{braha1998measurement}.
\begin{quote}
\textbf{In other words, the entropy function quantifies the complexity of the computational process needed to produce a configuration.}
\end{quote}
For example, in a Cartesian coordinate system (Fig. \ref{fig:spiral-cartesian}), the decision boundary of the double spiral problem  is nonlinear, while in a polar coordinate system, it is linear (Fig. \ref{fig:spiral-polar}.  Another very minimalistic example of representation and computation are two sides of the same coin, is the XOR function. Encoding inputs \& outputs on the domain $[0,1]$ requires a hidden layer of 3 neurons, while coding on $[-1,1]$ requires only two.  The representation thus clearly impacts the model entropy.

In this line of thought, constructs like \textit{structure} (organisation in layers), \textit{hierarchy} (connectivity only between layers), and drop-out are not essential to learning on themselves, but rather \textit{strategies} that have been implemented to lower the entropy.
  

\subsection{Minimizing model entropy leads to concept formation} 
According to Wikipedia, ``a concept is an abstract idea representing the fundamental characteristics of what it represents.  Concepts arise as abstractions or generalisations from experience or the result of a transformation of existing ideas." 
\citep{concept2016wikipedia}.  It thus makes sense to hypothesize that the \textit{minimization of entropy} accomplishes to capture the structural identities of concepts in the outside world (i.e. the training data).   For example, the model structure with optimal entropy to discriminate between hand-written numbers, is more probable to reflect the true nature of these concepts.  Empirical evidence supports this interpretation \citep{zahavy2016graying}, especially in the context of so-called \textit{one-shot generalization} \citep{rezende2016one}.

Finding or constructing such a entropy function is not an easy task and subject to further fundamental research.  More interesting, finding effective strategies to minimize this function, may prove even harder and deeper investigation into the complexity of models is needed. Today, often ad-hoc solutions are chosen like enforcing structural limitations (like Decision trees), pruning, regularization, weight sharing, kernel tricks, sparse representations etc.
In conclusion, 
\begin{quote}
\textbf{The minimization of the \textit{entropy function} is a necessary condition for concept formation.   In this process strategies like hierarchy, symmetry and other regularities can be exploited.}
\end{quote}

\section{Conclusion}
In this paper, I argue for a more holistic view on machine learning that takes a distance from the prevailing behaviourist perspective in which external validation is the major force to learning.  The work is centred around 5 conjectures:
\begin{enumerate}
\item Learning can not be explained in extrinsic terms only and data representation and model computation should be considered as two sides of the same coin. As a result the structural properties of both the model and data space are key to the modelling of higher abstractions.

\item Machine learning regards strategies that optimize performance, model structure and data modelling efficiency/likelihood. As a result, Machine learning is an intrinsically multi-objective process that trades off task performance for model complexity and data modelling efficiency/energy or model likelihood.

\item The definition of (machine) learning needs to reflect this vision: ``A computer program is said to learn from experience X if its performance at task in T, as measured by P, improves with experience X or when the model entropy S or its modelling efficiency or model likelihood E decreases with respect to the training data D associated with experience X."

\item An entropy function needs to be defined. This entropy function quantifies the complexity of the computational process needed to produce a configuration.

\item Minimizing model entropy leads to concept formation.  Hierarchy building, symmetry and exploitation of other regularities are strategies that drive optimization.
\end{enumerate}

Future work includes research on the entropy function and scientific and a quantitative validation of the postulated hypotheses on prototype examples. In a later phase, it should be investigated how the proposed extensions can improve existing Deep Learning techniques.

\bibliographystyle{named}
\bibliography{arxiv2017}

\end{document}